\documentclass{bmvc2k}

\title{\textbf{\methodsname}: Nested Multimodal 3D Representations via Matryoshka Representation Learning}

\addauthor{Márcus Lobo}{marcusvlc@usp.br}{1}
\addauthor{Vitor Matias}{vitorpmatias@usp.br}{1}
\addauthor{Jeová Farias}{j.farias@bowdoin.edu}{2}
\addauthor{Moacir Ponti}{moacir@icmc.usp.br}{1}

\addinstitution{
 Instituto de Ciências Matemáticas e de Computação \\
 Universidade de São Paulo \\
 São Carlos, São Paulo, BR
}
\addinstitution{
 Bowdoin College \\
 Brunswick, Maine, USA
}

\runninghead{Lobo \etal}{Nested Multimodal 3D Representations}

\def\etal{\emph{et al}\bmvaOneDot}

\usepackage{amssymb, amsfonts, amsthm}
\usepackage{booktabs}
\usepackage{multirow}
\usepackage{geometry}
\usepackage[table]{xcolor}
\usepackage{wrapfig}
\usepackage{graphicx}
\usepackage{capt-of}
\usepackage[dvipsnames]{xcolor}

\usepackage{soul}

\def\methodsname {\texttt{3D-MRL}\xspace}

\begin{document}

\maketitle

\begin{abstract}
Vision-Language Models align point clouds with image and text embeddings, enabling zero-shot recognition, retrieval, and open-vocabulary understanding of 3D shapes. Existing multimodal 3D pre-training methods produce fixed-dimensional embeddings, requiring separate models for different computational budgets.
We propose 3D Matryoshka Representation Learning (\methodsname), \textit{a multimodal 3D pre-training framework based on Matryoshka Representation Learning.} \methodsname learns nested 3D representations by aligning point clouds with frozen CLIP image and text embeddings while applying contrastive supervision across multiple embedding dimensions. The Matryoshka objective is applied only to the 3D encoder, allowing a single model to produce representations at different dimensionalities without retraining.
Experiments on the Objaverse-LVIS, ModelNet40, and ScanNet datasets show that \methodsname achieves competitive performance on zero-shot and few-shot 3D recognition tasks. In addition, the learned representations support retrieval across different embedding dimensions within a single model. On Objaverse-LVIS, \methodsname improves Top-1 accuracy from 46.8\% to 50.9\%. Retrieval experiments further show that different embedding dimensions yield varying levels of semantic and geometric specificity. Project page:~\url{https://usmarcv.github.io/3DMRL_page/}.

\end{abstract}

\section{Introduction}
\label{sec:intro}

Vision-Language Models (VLMs) for 3D perception align point clouds with images and text in a shared embedding space, enabling zero-shot classification, open-vocabulary retrieval, and cross-modal understanding of 3D shapes. Recent 3D multimodal learning methods such as ULIP~\cite{xue2023ulip} and OpenShape~\cite{liu2023openshape} achieve strong results by mapping 3D point clouds into CLIP's pre-aligned space. Despite their success, these methods learn and produce single fixed high-dimensional embeddings -- typically 512 to 1280 dimensions -- that must be stored and compared in full. For real-world deployment on edge devices, in large-scale retrieval pipelines, or under strict latency budgets, this one-size-fits-all approach is a critical bottleneck.

Matryoshka Representation Learning (MRL)~\cite{kusupati2022matryoshka} addresses this limitation by learning nested embeddings whose lower-dimensional prefixes remain useful for downstream tasks, while higher-dimensional prefixes progressively increase representational capacity.
A single MRL-trained model can be truncated to any prefix length, trading accuracy for efficiency without retraining. Thanks to the recent advances in building embeddings, MRL has proven effective in 2D vision~\cite{cai2025matryoshka}, language~\cite{li20242dmrl}, and, recently, 3D reconstruction with Gaussian primitives~\cite{guo2026mrlgs}. While MRL has demonstrated success in vision, language, retrieval, and graphics, its potential for multimodal 3D representation learning remains unexplored. 
Extending MRL to multimodal 3D is nontrivial because point cloud embeddings must preserve geometric structure~\cite{qi2017pointnet++, qi2017pointnet} while remaining aligned with frozen image and text representations. Unlike prior MRL settings in 2D or language, truncation in 3D can disproportionately remove local geometric cues critical for retrieval and recognition. Moreover, multi-scale contrastive supervision introduces competing alignment constraints across dimensions. Our formulation addresses these challenges via a logarithmic nesting schedule and parameter-free geometric truncation.

We introduce \textbf{3D Matryoshka Representation Learning (\methodsname)}, a framework that jointly aligns nested 3D embeddings with frozen CLIP image and text features at multiple dimensionalities.
~\autoref{fig:3d-mrl} shows an overview of the proposed framework. In contrast, while conventional 3D multimodal approaches that optimize a single embedding space apply a contrastive objective across multiple prefix lengths, \methodsname applies it across multiple prefix lengths. This encourages the 3D encoder to concentrate task-relevant information in the earliest dimensions while progressively refining the representation at larger dimensionalities.
At inference, the embedding is simply truncated to the available budget, tracing a good accuracy-efficiency frontier from a single model, without retraining or maintaining separate encoders for each operating point.

Our contributions are listed in threefold:
\begin{enumerate}
    \item We introduce \methodsname, the first framework to bring Matryoshka Representation Learning to 3D multimodal pre-training, learning a single encoder that produces high-quality embeddings at any prefix length.
    \item We achieve competitive performance on zero-shot and few-shot 3D recognition benchmarks, including Objaverse-LVIS, ModelNet, and ScanNet.
    \item We provide an analysis of the accuracy-efficiency trade-off enabled by nested 3D representations, showing that \methodsname degrades gracefully across a wide range of dimensionality budgets.
\end{enumerate}

Our results suggest that nested embedding structures provide an effective mechanism for organizing 3D multimodal representations across multiple levels of capacity, enabling scalable and budget-aware 3D understanding from a single model.

\vspace{-0.4cm}
\section{Related Work}
\label{sec:related}

\vspace{0.15cm}
\noindent \textbf{3D Multimodal Learning.} Learning representations from 3D point clouds has been tackled through two complementary paradigms: projecting point clouds onto regular grids (voxels or multi-view images) for processing with 2D/3D convolutions~\cite{choy20194dsparseconv, shi2020pv}, and directly modeling unstructured point sets with point-centric architectures such as PointNet~\cite{qi2017pointnet, qi2017pointnet++}, PointTransformer~\cite{wu2022pointtransformer}, or masked autoencoders~\cite{yu2022point, zhang2022pointmae}.
Independently, VLMs led by CLIP~\cite{radford2021learning} established contrastive learning as a powerful mechanism for cross-modal alignment, mapping images and text into a shared embedding space. 
This paradigm was extended to 3D: ULIP~\cite{xue2023ulip} and OpenShape~\cite{liu2023openshape} align point-cloud features with pretrained image and text embeddings, while Uni3D~\cite{zhou2024uni3d} scales this formulation with a vision transformer-based 3D encoder initialized from 2D models.
Succeeding methods improve different aspects of this alignment. TAMM~\cite{zhang2024tamm} and MixCon3D~\cite{gao2024sculpting} strengthen multimodal 3D representation learning, OpenDlign~\cite{mao2024opendlign} improves visual supervision through depth-aligned synthesized images, and MRD~\cite{wang2024multimrd} distills intra- and cross-modal relational structure to preserve the semantic organization of the pretrained vision-language space. 
More recent approaches further specialize this paradigm: TriCLIP-3D~\cite{li2025triclip} develops tri-modal alignment for 3D visual grounding, while DuoduoCLIP~\cite{lee2025duoduo} and Adaptive CLIP~\cite{song2025adaptive} explore efficient multi-view modeling and retrieval-oriented 3D--CLIP alignment, respectively. Cross-modal methods combining point clouds and multi-view imagery have also continued to improve 3D representation learning~\cite{zhou2025crossmodal}.
Despite advances in model scale, supervision, and cross-modal alignment, these methods commonly optimize a single fixed-dimensional embedding. As a result, deployment under different memory or latency budgets requires either training separate models or applying post-hoc dimensionality reduction, which may degrade the learned semantic structure. Our work instead learns a nested hierarchy of 3D representations during multimodal contrastive pre-training, enabling a single encoder to operate across multiple embedding budgets without retraining.

\vspace{0.15cm}
\noindent \textbf{Matryoshka Representation Learning.} Matryoshka Representation Learning (MRL)~\cite{kusupati2022matryoshka} introduced a simple but consequential idea: a single representation vector can be trained such that every prefix, the first dimensions $d$ for $d \in \{d_1, \dots, \mathcal{D}\}$ is independently useful for downstream tasks. This nested structure provides a built-in accuracy-efficiency knob without
retraining or model switching.
The principle has since been extended along several axes. 2D-MRL~\cite{li20242dmrl} nests representations across both embedding dimensions and transformer layers, showing the idea generalizes beyond a
single feature vector. 
MetaEmbed~\cite{xiao2025metaembed} adapts the Matryoshka principle to multi-vector retrieval, learning nested groups of meta tokens that enable test-time scaling of retrieval quality against index size. In the same spirit, Matryoshka Multimodal Models (M$^3$)~\cite{cai2025matryoshka} learn nested sets of visual tokens for vision-language generation, allowing the token budget per image to be adjusted at inference time.
Also, Matryoshka Gaussian Splatting~\cite{guo2026mrlgs} demonstrates that the nesting principle extends beyond embeddings to geometric primitives, learning an ordered set of 3D Gaussians whose prefixes yield coherent scene reconstructions at any rendering budget.
Recent work has also revisited the compression mechanism itself.
CSR~\cite{wen2025beyondcsr} replaces prefix truncation with sparse representations, retaining only the most informative embedding dimensions while preserving semantic structure through reconstruction and contrastive objectives. SMEC~\cite{zhang2025smec} improves Matryoshka compression through sequential dimensional reduction and adaptive dimension selection, reducing interference across embedding sizes and preserving similarity structure.

Despite this growing body of work across vision, language, retrieval, and graphics, the Matryoshka principle remains entirely unexplored in 3D multimodal representation learning -- none of the recent 3D-CLIP methods \cite{xue2023ulip, liu2023openshape, li2025triclip, lee2025duoduo} incorporate or evaluate nested or multi-granularity representations. Our work fills this gap. \methodsname trains the encoder to produce a nested hierarchy of representations. This is conceptually distinct from applying MRL post-hoc: by coupling the nested structure with cross-modal contrastive pre-training from the start, we ensure that dimensionality reduction becomes a property of the learned space.

\vspace{-0.4cm}
\section{3D Matryoshka Representation Learning}
\label{sec:method}

\begin{wrapfigure}{l}{0.45\textwidth}
    \vspace{-0.5cm}
    \includegraphics[width=\linewidth]{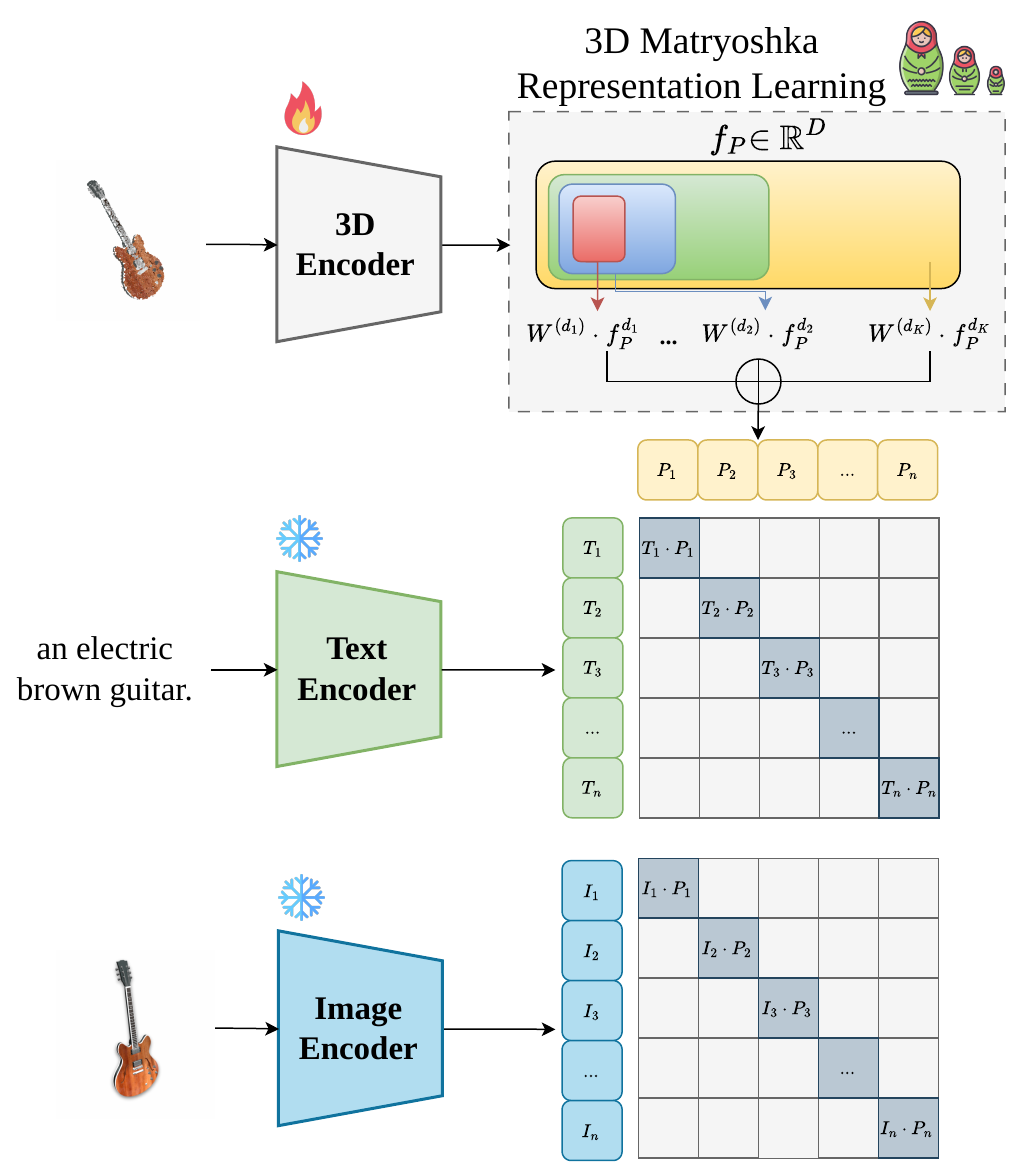}
    \vspace{-0.8cm}
    \caption{\textbf{Overview of~\methodsname.} \methodsname is a framework that aligns nested 3D embeddings with multiple dimensionalities with multimodal image and text for 3D representation learning. \methodsname shows competitive performance on a wide range of benchmarks.}
    \label{fig:3d-mrl}
\end{wrapfigure}

\methodsname learns nested 3D representations by aligning point clouds with frozen image and text embeddings. Unlike prior multimodal 3D methods that optimize a single embedding space, \methodsname applies contrastive supervision across multiple embedding prefixes generated by the 3D encoder. This allows a single model to produce representations that remain effective across a range of dimensionality budgets. ~\autoref{fig:3d-mrl} provides an overview of the proposed framework. The Matryoshka learning strategy is applied only to the 3D encoder, whereas the image and text encoders remain frozen and provide fixed multimodal targets throughout training.

Our central hypothesis is that semantic abstraction emerges progressively along the dimensions of the learned 3D representation. Lower-dimensional prefixes are encouraged to retain broad category-level semantics under severe compression, while additional dimensions progressively refine geometric and structural information. To encourage this behavior, \methodsname extends multimodal contrastive learning with a Matryoshka design objective that jointly aligns nested 3D embeddings with frozen image and text representations across multiple embedding scales.

\vspace{0.15cm}
\noindent \textbf{Problem Formulation.}
Let $\mathcal{S} = \{(P_i, I_i, T_i)\}_{i=1}^{n}$ be a dataset of $n$ triplets, where $P_i$ is a 3D point cloud, $I_i$ is its corresponding 2D rendered image from a 3D object, and $T_i$ is an associated textual description generated by a vision-language model such as BLIP~\cite{li2023blip}.

We define a new 3D encoder $E_P: \mathcal{P} \to \mathbb{R}^{D}$ that maps point clouds to $D$-dimensional embeddings, and use frozen CLIP encoders $E_I: \mathcal{I}(I) \to \mathbb{R}^{D}$ and $E_T: \mathcal{T}(T) \to \mathbb{R}^{D}$ for images and text, respectively.
For a given triplet from a dataset $(P_i, I_i, T_i)$, w.l.o.g. we obtain three embeddings, 
\[
f_{P} = E_P(P), \quad
f_{I} = E_I(I), \quad
f_{T} = E_T(T),
\]
where all embeddings are $\ell_2$-normalized to lie on the unit hypersphere $\mathcal{S}^{D-1}$, $E_I$ and $E_T$ are frozen CLIP encoders~\cite{radford2021learning, cherti2023reproducible}, and $E_P$ is a trainable 3D encoder.  

\vspace{0.15cm}
\noindent \textbf{Nested Representation Structure.}
The core principle of \methodsname is that the 3D encoder $E_P$ should produce representations whose prefixes form a hierarchy of valid embeddings, ranging from broad semantic abstractions to increasingly fine-grained geometric and structural representations. Formally, let
\[
\mathcal{D} = \{d_1, d_2, \dots, d_K\}, \quad
1 \leq d_1 < \dots < d_K = \mathcal{D}
\]
be an ordered set of embedding dimensionalities. To align with the frozen CLIP encoders, we set the maximum dimension to $\mathcal{D}=1280$ and the base dimension to $d_1=10$. Following the Matryoshka Representation Learning design~\cite{kusupati2022matryoshka}, we construct $\mathcal{D}$ as a logarithmic schedule that scales geometrically from this base, resulting in $\mathcal{D} = \{10, 20, 40, \dots, 1280\}$. This allocates denser supervision to low-dimensional embeddings while progressively increasing representational capacity. Intuitively, low-dimensional prefixes are encouraged to retain only the most informative semantic features, whereas additional dimensions progressively encode finer geometric details.
Optimizing over these $K=8$ nested dimensions provides dense coverage across multiple levels of granularity, from compact semantic embeddings to high-capacity representations, while introducing only minimal optimization overhead.

For any target dimension $d_k \in \mathcal{D}$, the \emph{nested sub-embedding} is obtained by isolating the vector prefix containing the first $d_k$ continuous coordinates of $f_{P}$, denoted by the slicing operator $[1\!:\!d_k]$:
$
f^{d_k}_{P} = f_{P}[1\!:\!d_k] \; \in \; \mathbb{R}^{d_k}.
$
This slice is subsequently re-normalized to unit norm to preserve directional consistency. The nesting constraint guarantees that coarser representations are strict prefixes of finer ones ($f^{d_a}_{P} \subset f^{d_b}_{P}$ for $d_a < d_b$).

\vspace{0.15cm}
\noindent \textbf{Multi-Scale Contrastive Alignment.}
To learn representations that are discriminative at every scale, we apply a contrastive objective independently at each dimensionality $d_k$. In practice, because the image and text representations are derived from frozen CLIP encoders with a fixed maximum dimension $\mathcal{D}$ (i.e., 1280), they cannot dynamically adapt their internal structure. Therefore, to evaluate the cross-modal similarity at a reduced capacity $d_k$, we apply the identical prefix slicing operation directly to the target modalities, yielding $f^{d_k}_{I}$ and $f^{ d_k}_{T}$, followed by unit $\ell_2$-renormalization.

To enforce a structured, discriminative hierarchy, \methodsname optimizes this multimodal alignment using a dedicated linear classifier (or projection head) parameterized by $\mathbf{W}^{(d_k)} \in \mathbb{R}^{d_k \times D}$ for each discrete scale. Given two modalities $M_1$ (3D to 2D images) and $M_2$ (3D to texts) with batch embeddings truncated to scale $d_k$, the InfoNCE loss~\cite{oord2018representation} with temperature $\tau > 0$ is formulated as:
\begin{equation}
\label{eq:contrast}
\mathcal{L}_{\text{NCE}}(f^{d_k}_{M_1}, f^{d_k}_{M_2}) =
-\frac{1}{n}\sum_{i=1}^{n}
\log\frac{
\exp\left( \langle 
f^{ d_k}_{M_1}, 
f^{d_k}_{M_2} \rangle / \tau \right)
}{
\sum_{j} \exp\left( \langle f^{d_k}_{ M_1}, f^{d_k}_{M_2,j} \rangle / \tau \right)
},
\end{equation}
where $\langle\cdot,\cdot\rangle$ denotes cosine similarity, $f^{d_k}_{M_2}$ represents the positive augmented sample originating from the same source sample as the anchor, and the index $j$ iterates over all augmented samples in the batch (both the positive sample and all negative samples).

We jointly optimize the parameters $\theta_P$ of the 3D encoder $E_P$ and the set of scale-specific linear classifiers $\{\mathbf{W}^{(d_k)}\}_{k=1}^K$ by minimizing the aggregated multi-scale loss:
\begin{equation}
\label{eq:3dmrl}
\min_{\{\mathbf{W}^{(d_k)}\}_{k=1}^K, \, \theta_P} \sum_{k=1}^{K} \lambda_k \cdot \left[ \mathcal{L}_{\text{NCE}}\left(\mathbf{W}^{(d_k)} \cdot f^{d_k}_{P}, \; f^{d_k}_{I}\right) + \mathcal{L}_{\text{NCE}}\left(\mathbf{W}^{(d_k)} \cdot f^{d_k}_{P}, \; f^{d_k}_{T}\right) \right],
\end{equation}
\noindent where $\lambda_k \geq 0$ is a per-scale weighting coefficient (set to $\lambda_k = 1$ by default). Crucially, despite explicitly optimizing only for the $K = 8$ chosen log-spaced granularities ($O(\log(D))$ dimensions), the nesting structure naturally induces accurate representation spaces that smoothly interpolate for intermediate dimensions falling between the designated scales.

\vspace{0.15cm}
\noindent \textbf{Efficient vs. Non-Efficient multimodal MRL.}
The scale-specific linear transformations $\mathbf{W}^{(d_k)} \in \mathbb{R}^{d_k \times d_k}$ are used only during pre-training to facilitate alignment across embedding scales. Unlike conventional multi-scale approaches that require independent projection heads from the full-dimensional representation space ($\mathbb{R}^{D}\rightarrow\mathbb{R}^{d_k}$), our formulation relies primarily on parameter-free prefix truncation and applies lightweight transformations only within each nested subspace.
This design substantially reduces parameter overhead while preserving a shared hierarchical representation across all dimensions. \emph{For more details, see Section~\ref{subsec:ablations} in our ablation study}.

\vspace{0.15cm}
\noindent \textbf{Properties of the Learned Representation.}
The objective in Eq.~\eqref{eq:3dmrl} explicitly supervises every prefix
$f_P^{d_k}$ of the representation. As a result, the learned embedding exhibits two important properties.


\noindent
\textit{(i) Nested semantic consistency.}
For every scale pair $d_i < d_j$, the smaller representation is contained within the larger one,
$
f_P^{d_i}
=
\left(
f_P^{d_j}
\right)_{1:d_i},
$
\noindent ensuring that semantic information learned at lower dimensions is preserved across all larger embedding scales.


\noindent
\textit{(ii) Progressive representational refinement.}
As dimensionality increases, the representation gains additional degrees of freedom while preserving lower-dimensional semantics. 
\begin{wrapfigure}{r}{0.35\textwidth}
    \vspace{0.25cm}
    \centering
    \includegraphics[width=1\linewidth]{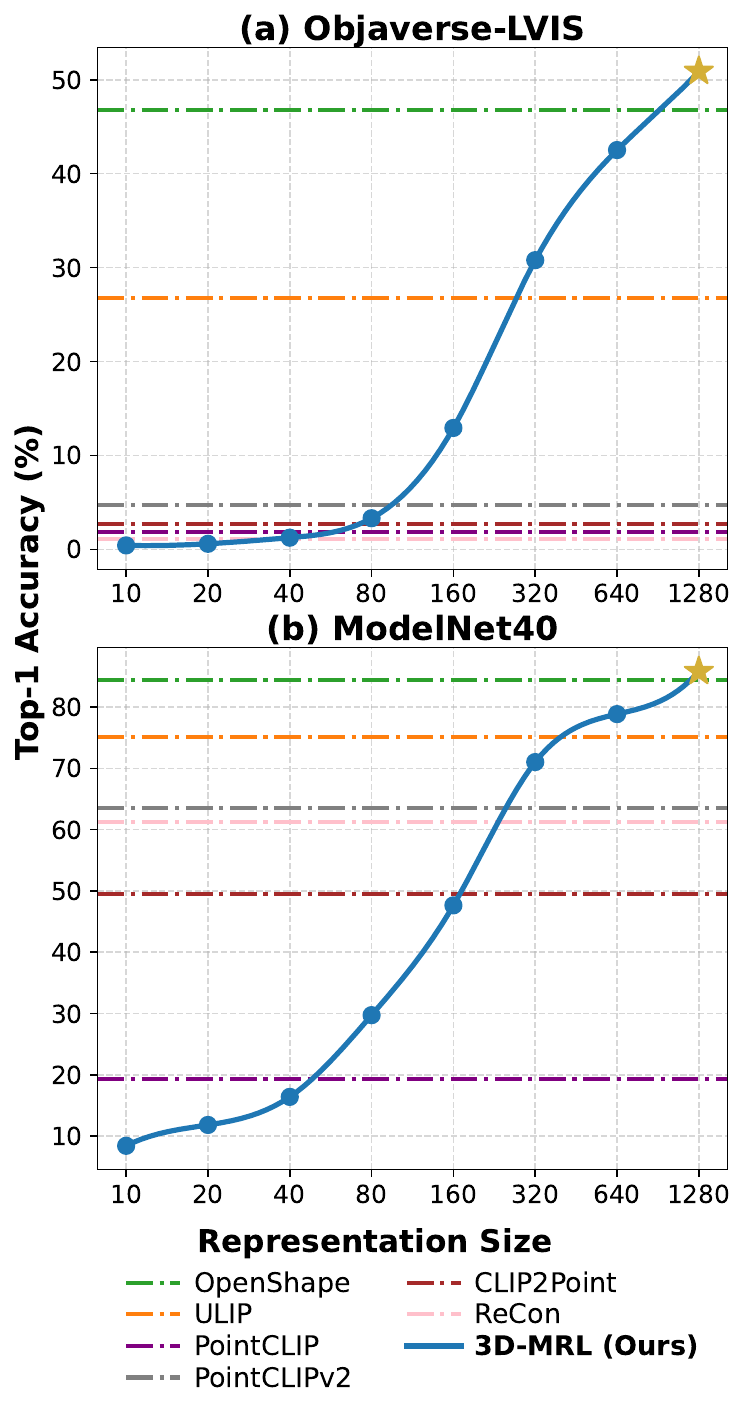}
    \vspace{-0.8cm}
    \caption{Zero-shot 3D shape classification on the Objaverse-LVIS~\cite{deitke2023objaverse} (1,156 categories) and ModelNet40~\cite{wu20153dmodelnet}.}
    \label{fig:teaser_zsl}
    \vspace{-1.2cm}
\end{wrapfigure}
Formally,
$
\dim(f_P^{d_i})
<
\dim(f_P^{d_j}),$
$
d_i < d_j,
$ allowing larger prefixes to encode information unavailable to smaller embeddings. Consequently, the representation can progressively capture finer semantic and geometric distinctions without discarding information learned at lower dimensions.

\vspace{0.15cm}
\noindent \textbf{Application in Downstream Tasks.}
For zero-shot 3D classification, we aggregate similarity scores across all $K$ scales between the nested 3D embedding and category text embeddings:
\[
s(c \mid P_i) = \sum_{k=1}^{K} \alpha_k \;
\langle f^{d_k}_{P}, \; f^{d_k, c}_{T} \rangle,
\]
where $f_T^{c}$ is the frozen CLIP text embedding for category $c$, and $\alpha_k$ are aggregation weights (we use uniform weighting by default). The predicted class is $\hat{c} = \arg\max_c \, s(c \mid P)$.
For few-shot classification, we follow the standard $K$-way $N$-shot protocol. For each trial, a linear classifier is trained on top of the frozen nested embeddings $f^{d_k}_{P}$ at each scale $d_k$, while the 3D encoder $E_P$ remains fixed. By sweeping $d_k \in \mathcal{D}$ and reporting accuracy at each dimensionality, we characterize the full accuracy-compactness trade-off from a single pre-trained model, with no retraining or architecture changes required.
~\autoref{fig:teaser_zsl} summarizes the performance of~\methodsname.

\vspace{-0.4cm}
\section{Experiments}
\label{sec:experiments}
In this section, we first introduce our experimental setup in Section~\ref{subsec:setup}. Then, we show the performance of \methodsname on downstream tasks such as zero-shot 3D shape classification (Section~\ref{subsec:zero-shot-cls}) and the matched-budget compression and efficiency (Section~\ref{subsec:matched-budget-efficiency}),  few-shot linear probing 3D classification (Section~\ref{subsec:few-shot}), and real-world recognition (Section~\ref{subsec:real-world}). Additionally, our~\methodsname supports cross-modal applications such as point cloud and text-to-3D shape retrieval (Section~\ref{subsec:cross-modal}). Finally, we provide some ablations in Section~\ref{subsec:ablations}.

\subsection{Experimental Setup}
\label{subsec:setup}

\vspace{0.15cm}
\noindent \textbf{Pre-training Datasets.}
To keep the experimental settings consistent with other methods, we follow OpenShape~\cite{liu2023openshape} for a fair comparison. Our \methodsname is pre-trained on the triplets generated from four datasets (denoted as ``Ensembled''): ShapeNetCore~\cite{chang2015shapenet},  3D-FUTURE~\cite{future20213d}, ABO~\cite{collins2022abo}, and Objaverse~\cite{deitke2023objaverse}. 
``ShapeNet'' is a triplet set derived from the ShapeNetCore dataset, containing 52,470 3D shapes, corresponding images, and text; Ensembled (no LVIS) is a set of 829,460 triplets from the above datasets except Objaverse-LVIS~\cite{gupta2019lvis}; ``Ensembled'' denotes the triplet set comprising data from all four datasets (mentioned above), containing 875,665 3D shapes and their associated images and text.

The point cloud is obtained by sampling 10k points from the mesh surface, and the colors are interpolated from the mesh textures. The images are rendered from 12 preset camera poses that uniformly cover the entire object. The paired texts are generated by BLIP~\cite{li2022blip, li2023blip} and Azure services with GPT-4~\cite{achiam2023gpt} to filter out noisy text. 

\vspace{0.15cm}
\noindent \textbf{Downstream Datasets.}
Our \methodsname is evaluated on the following datasets: ModelNet40 \cite{wu20153dmodelnet} is a synthetic dataset with 3D CAD models, including 9,843 samples and 2,468 testing samples, distributed across 40 3D categories; ScanObjectNN~\cite{scanobj2019} is a dataset composed of 3D objects acquired throught real-world scanning techniques, encompassing a total of 2,902 objects that are categorized into 15 distinct categories; Objaverse-LVIS~\cite{gupta2019lvis}, is an annotated subset from Objaverse~\cite{deitke2023objaverse}, incorporates a corpus of 46,832 shapes originating from 1,156 categories in LVIS dataset~\cite{gupta2019lvis}.

\vspace{0.15cm}
\noindent \textbf{Downstream Tasks.}
We conduct experiments on three tasks, including zero-shot 3D shape classification, few-shot linear probing 3D classification, and real-world recognition, to de\-monstrate the advantages of our \methodsname. \emph{Other implementation details regarding pre-train\-ing and evaluation are introduced in the supplementary material}.

\subsection{Zero-Shot 3D Shape Classification}
\label{subsec:zero-shot-cls}
We first evaluate \methodsname on the zero-shot shape classification task, a key metric for assessing the quality of the learned 3D representation, which requires the representations to be directly applicable to datasets where the model has never been explicitly supervised. Without any further tuning, the 3D representations are compared with text and image embeddings of categories to predict the classes of 3D shapes.
To make a fair comparison and keep consistency with prior work, we adopt the same setting as OpenShape~\cite{liu2023openshape} (i.e., $\mathcal{D} = 1280$) on three benchmarks: Objaverse-LVIS~\cite{gupta2019lvis}, ModelNet40~\cite{wu20153dmodelnet}, and ScanObjectNN(OBJ-ONLY)~\cite{scanobj2019}. We compare \methodsname with existing zero-shot approaches, including PointCLIP~\cite{zhang2022pointclip}, PointCLIPv2~\cite{zhu2023pointclipv2}, CLIP2Point~\cite{huang2023clip2point}, ReCon~\cite{qi2023contrast}, ULIP~\cite{xue2023ulip} and OpenShape~\cite{liu2023openshape}. Among them, PointCLIP~\cite{zhang2022pointclip} and PointCLIPv2~\cite{zhu2023pointclipv2} project point clouds into 2D images and directly use 2D CLIP for inference, whereas the other methods leverage CLIP embedding spaces for alignment without matryoshka embeddings and require 3D shapes for training.

\begin{table}[h]
\centering
\resizebox{\textwidth}{!}{%
\begin{tabular}{c | c | ccc | ccc | ccc}
\toprule
\multirow{2}{*}{Method} & \multirow{2}{*}{\shortstack{Pre-Training\\Dataset}} & \multicolumn{3}{c|}{Objaverse-LVIS~\cite{deitke2023objaverse}} & \multicolumn{3}{c|}{ModelNet40~\cite{wu20153dmodelnet}} & \multicolumn{3}{c}{ScanObjectNN~\cite{scanobj2019}} \\
\cmidrule{3-11}
 & & Top1 & Top3 & Top5 & Top1 & Top3 & Top5 & Top1 & Top3 & Top5 \\
\midrule
PointCLIP~\cite{zhang2022pointclip} & - & 1.9 & 4.1 & 5.8 & 19.3 & 28.6 & 34.8 & 10.5 & 20.8 & 30.6 \\
PointCLIPv2~\cite{zhu2023pointclipv2} & - & 4.7 & 9.5 & 12.9 & 63.6 & 77.9 & 85.0 & 42.2 & 63.3 & 74.5 \\
\midrule
CLIP2Point~\cite{huang2023clip2point} & \multirow{7}{*}{ShapeNet} & 2.7 & 5.8 & 7.9 & 49.5 & 71.3 & 81.2 & 25.5 & 44.6 & 59.4 \\
ReCon~\cite{qi2023contrast} & & 1.1 & 2.7 & 3.7 & 61.2 & 73.9 & 78.1 & 42.3 & 62.5 & 75.6 \\
ULIP-PointBERT~\cite{xue2023ulip} & & 6.2 & 13.6 & 17.9 & 60.4 & 79.0 & 84.4 & 51.5 & 71.1 & 80.2 \\
OpenShape-SparseConv~\cite{liu2023openshape} & & 11.6 & 21.8 & 27.1 & 72.9 & 87.2 & \textbf{93.0} & \textbf{52.7} & 72.7 & \textbf{83.6} \\
OpenShape-PointBERT~\cite{liu2023openshape} & & 10.8 & 20.2 & 25.0 & 70.3 & 86.9 & 91.3 & 51.3 & 69.4 & 78.4 \\
\methodsname-SparseConv (Ours)~\cite{liu2023openshape} & &  \textbf{12.9} & \textbf{23.7} &\textbf{29.4} & \textbf{73.1} & \textbf{88.4} & 91.8 & 49.6 & \textbf{73.0} & 82.1  \\
\methodsname-PointBERT (Ours)~\cite{liu2023openshape} & &  12.4 & 22.1 & 27.2 & 71.8 & 86.8 & 91.3 & 51.2 & 70.9 & 80.3   \\ 

\midrule

ULIP-PointBERT~\cite{xue2023ulip} & \multirow{5}{*}{\shortstack{Ensembled\\(no LVIS)}} & 21.4 & 38.1 & 46.0 & 71.4 & 84.4 & 89.2 & 46.0 & 66.1 & 76.4 \\
OpenShape-SparseConv~\cite{liu2023openshape} & & 37.0 & 58.4 & 66.9 & 82.6 & 95.0 & 97.5 & \textbf{54.9} & 76.8 & \textbf{87.0} \\
OpenShape-PointBERT~\cite{liu2023openshape} & & 39.1 & 60.8 & 68.9 & 85.3 & \textbf{96.2} & 97.4 & 47.2 & 72.4 & 84.7 \\
\methodsname-SparseConv (Ours) & & 39.2 & 60.4 & 68.8 & 83.4 & 95.6 & \textbf{97.6} & 53.9 & \textbf{76.8} & 85.7 \\
\methodsname-PointBERT (Ours) & & \textbf{40.3} & \textbf{61.1} & \textbf{69.2} & \textbf{85.5} & 95.7 & 97.2 & 50.4 & 75.9 & 84.8 \\

\midrule

ULIP-PointBERT~\cite{xue2023ulip} & \multirow{5}{*}{Ensembled} & 26.8 & 44.8 & 52.6 & 75.1 & 88.1 & 93.2 & 51.6 & 72.5 & 82.3 \\
OpenShape-SparseConv~\cite{liu2023openshape} & & 43.4 & 64.8 & 72.4 & 83.4 & 95.6 & 97.8 & \textbf{56.7} & 78.9 & 88.6 \\
OpenShape-PointBERT~\cite{liu2023openshape} & & 46.8 & 69.1 & 77.0 & 84.4 & \textbf{96.5} & 98.0 & 52.2 & \textbf{79.7} & \textbf{88.7} \\
\methodsname-SparseConv (Ours) & &  42.7 & 64.4 & 72.5 & 84.7 & 95.9 & \textbf{98.1} & 54.2 & 77.9 & 87.0 \\
\methodsname-PointBERT (Ours) & & \textbf{50.9} &\textbf{ 72.6} & \textbf{79.6}
 & \textbf{85.8} & 96.3 & 97.8 & 49.6 & 73.3 & 84.3 \\
\bottomrule
\end{tabular}%
}
\caption{Zero-shot classification on Objaverse-LVIS~\cite{deitke2023objaverse}, ModelNet40~\cite{wu20153dmodelnet}, and ScanObjectNN~\cite{scanobj2019}. The performance gains from~\methodsname are more significant on the most challenging long-tailed Objaverse-LVIS dataset. In bold, we mark the best results per pre-training method.}
\label{tab:zero-shot-results}

\end{table}

The quantitative comparisons are summarized in~\autoref{tab:zero-shot-results}. We observe that \methodsname leverages matryoshka multimodal pre-training to consistently outperform PointCLIP~\cite{zhang2022pointclip} and PointCLIPv2~\cite{zhu2023pointclipv2} by a large margin. In this scenario,~\methodsname consistently outperforms prior approaches when trained only on ShapeNet. 
For example, PointBERT~\cite{yu2022point} pre-trained by our \methodsname on the Ensembled dataset surpasses ULIP~\cite{xue2023ulip} and OpenShape~\cite{liu2023openshape} by margins of $+24.1\%$ and $+4.1\%$ in Top-1 accuracy on the long-tailed Objaverse-LVIS benchmark, which demonstrates the effectiveness of the multimodal MRL pre-training scheme. 
As we can see, \methodsname achieves its largest performance gains on Objaverse-LVIS, which contains diversity and long-tailed object categories. We hypothesize that the nested embedding structure gains from open-vocabulary recognition scenarios, where coarse semantic information and fine-grained geometric cues must coexist within the same representation space. The lower-dimensional embeddings appear to capture broad semantic concepts across large-scale objects, while higher-dimensional subspaces progressively specialize toward specific category details. \emph{Additional results across embedding dimensions are provided in the supplementary material}.

As for ModelNet40, \methodsname achieves an $85.8\%$ top-1 accuracy, surpassing OpenShape by $+1.4\%$ for the PointBERT model.
On ScanObjectNN, which contains challenging real-world scans with noise and occlusions, the performance of \methodsname remains comparatively fair. This highlights the difficulty of sim-to-real transfer in zero-shot 3D scenarios, particularly when representations are primarily learned from synthetic or rendered data. However, \methodsname with the SparseConv achieves competitive performance with OpenShape under the same experimental setting.

\vspace{-0.4cm}
\subsection{Matched-Budget Compression and Efficiency}
\label{subsec:matched-budget-efficiency}

To quantify the accuracy--efficiency trade-off, we compare \methodsname and OpenShape under matched embedding budgets on  Objaverse-LVIS, as shown in ~\autoref{tab:matched_budget}.
Direct \methodsname prefixes remain competitive across dimensions and outperform OpenShape at larger budgets (42.5\% vs. 40.2\% at 640D and 50.9\% vs. 46.6\% at 1280D). Moreover, \methodsname is substantially more robust to post-hoc PCA compression, reaching 27.3/43.3/50.0\% Top-1 accuracy at 40/80/160D, compared with 14.7/29.5/40.8\% for OpenShape. 
In~\autoref{tab:retrieval_efficiency}, these compact representations also provide direct retrieval benefits: using 320D instead of 1280D reduces storage by 4$\times$ (225.61 to 56.40 MB) and CPU retrieval latency (Lat.) by 4.1$\times$ (5.11 to 1.26 ms), demonstrating that a single \methodsname encoder can adapt to different accuracy--storage--latency budgets without retraining. \emph{More details in the supplementary material.}

\begin{table*}[h]
\centering

\begin{minipage}[t]{0.55\textwidth}
\centering
\setlength{\tabcolsep}{3pt}
\renewcommand{\arraystretch}{0.9}
\resizebox{\linewidth}{!}{%
\begin{tabular}{lcccccccc}
\toprule
Method $\backslash$ $\mathcal{D}$
& 10 & 20 & 40 & 80 & 160 & 320 & 640 & 1280 \\
\midrule
\methodsname
& 0.4 & 0.6 & 1.3 & 3.3 & 12.9 & 30.8 & 42.5 & \textbf{50.9} \\
\methodsname $+$ PCA
& \textbf{2.4} & \textbf{9.6} & \textbf{27.3} & \textbf{43.3}
& \textbf{50.0} & \textbf{51.0} & \textbf{50.6} & 50.5 \\
OpenShape truncated.
& 0.4 & 0.6 & 1.9 & 4.9 & 15.1 & 31.2 & 40.2 & 46.6 \\
OpenShape $+$ PCA
& 2.3 & 6.0 & 14.7 & 29.5 & 40.8 & 44.8 & 45.1 & 45.1 \\
\bottomrule
\end{tabular}%
}
\captionof{table}{Matched-budget zero-shot classification on a pretrained Ensembled dataset.}
\label{tab:matched_budget}
\end{minipage}
\hfill
\begin{minipage}[t]{0.40\textwidth}
\centering
\setlength{\tabcolsep}{3pt}
\renewcommand{\arraystretch}{0.7}
\resizebox{\linewidth}{!}{%
\begin{tabular}{cccc}
\toprule
$\mathcal{D}$ & Storage (MB) & Lat. (ms) & QPS \\
\midrule
160  & 28.20  & 0.477 & 23,122.9 \\
320  & 56.40  & 1.259 & 14,827.4 \\
640  & 112.81 & 2.458 & 9,559.7 \\
1280 & 225.61 & 5.107 & 5,336.6 \\
\bottomrule
\end{tabular}%
}
\captionof{table}{FAISS retrieval efficiency. QPS means queries per second.}
\label{tab:retrieval_efficiency}
\end{minipage}

\end{table*}

\vspace{-0.8cm}
\subsection{Few-Shot Linear Probing 3D Classification}
\label{subsec:few-shot}
Following the previous evaluation, we perform few-shot learning classification experiments over ModelNet~\cite{wu20153dmodelnet} to assess \methodsname under low-data regimes across nested embedding dimensions. Here, we extend our model with an additional linear classification layer with a matryoshka regime and train only this layer, rather than fine-tuning the entire model. 
Following previous work~\cite{yu2022point, zhang2022point, zhang2024tamm} we adopt the ``$K-$way $N-$shot'' setup, wherein we select $K$ classes at random and sample $(N + 20)$ instances per class.  We train on a support set of $K \times X$ samples, while evaluation uses a query set of the remaining 20 instances per class. 
We assess our model under four distinct scenarios: ``5-way 10-shot'', ``5-way 20-shot'', ``10-way 10-shot'' and ``10-way 20-shot''. For each scenario, we run 10 separate trials and report the mean performance and standard deviation across these trials, using only the highest-dimensional embedding ($d_8$=$1280$). 
\textit{For more details across the granularity dimensions for this task, see the supplementary materials.}
\begin{table}[h]
\centering

\begin{tabular}{lc|cccc}
\toprule
\multirow{2}{*}{\begin{tabular}[c]{@{}l@{}}Pre-Training\\ Dataset\end{tabular}} & \multirow{2}{*}{Method} & \multicolumn{2}{c}{5-way} & \multicolumn{2}{c}{10-way} \\
 &  & 10-shot & 20-shot & 10-shot & 20-shot \\ \midrule
\multirow{3}{*}{ShapeNet} & ULIP~\cite{xue2023ulip} & 94.4$\pm$3.7 & 93.2$\pm$4.2 & 86.6$\pm$5.3 & 90.6$\pm$5.2 \\
 & OpenShape~\cite{liu2023openshape} & 95.3$\pm$2.6 & 97.9$\pm$3.9 & 89.2$\pm$5.1 & 92.9$\pm$3.9 \\
 & \methodsname$^*$ (Ours) & 95.2$\pm$2.7 &  96.1$\pm$2.4 & 88.2$\pm$6.5  & 91.1$\pm$4.6  \\ 
\midrule
\multirow{2}{*}{Ensembled} & OpenShape~\cite{liu2023openshape} & \textbf{96.1$\pm$2.7} & 95.7$\pm$2.5 & 89.1$\pm$4.6 & 91.8$\pm$3.7 \\
 & \methodsname~$^*$ (Ours) & 94.8$\pm$2.8 & \textbf{97.9$\pm$0.9} &  \textbf{89.7$\pm$6.0} & \textbf{93.2$\pm$3.1} \\ 
 
\bottomrule
\end{tabular}

\scriptsize{$^*$Results using PointBERT~\cite{devlin2019bert} as 3D encoder, pre-trained on the Ensembled dataset.}

\caption{Few-shot linear probing classification results on ModelNet40. We report the average accuracy and standard deviation of 10 independent experiments. Our \methodsname remains competitive across few-shot settings and improves over OpenShape in three of four Ensembled configurations.}
\label{tab:few-shot-results}
\end{table}

As illustrated in~\autoref{tab:few-shot-results}, \methodsname achieves competitive performance across multiple few-shot evaluation settings, particularly when pre-trained on the Ensembled dataset using PointBERT. In particular, \methodsname surpasses OpenShape~\cite{liu2023openshape} by margins of $2.2\%$, $0.6\%$, and $1.4\%$ on the ``5-way 20-shot'', ``10-way 10-shot'', and ``10-way 20-shot'' settings, respectively.
Notably, the improvements become more evident in the more challenging ``10-way'' evaluation settings, suggesting that the hierarchical embedding structure becomes increasingly beneficial as the semantic complexity of the classification task grows.

The few-shot performance suggests that the nested representation already organizes semantic concepts in a linearly separable manner prior to downstream adaptation. Lower-dimensional subspaces appear to preserve robust category-level semantic information, while higher-dimensional embeddings progressively refine fine-grained geometric details. As a result, lightweight linear classifiers can effectively adapt under limited supervision without requiring full model fine-tuning.
These observations indicate that the proposed multi-granularity representation promotes transferable semantic structures rather than dataset-spe\-cific memorization, enabling \methodsname to generalize effectively under low-data regimes.

\vspace{-0.3cm}
\subsection{Real-World Recognition}
\label{subsec:real-world}

To evaluate the capability of \methodsname to understand real-world 3D shapes and scenes, we follow CLIP$^2$~\cite{zeng2023clip2} and conduct experiments on the ScanNet~\cite{dai2017scannet} dataset under the zero-shot recognition setting. In this task, the model aims to classify object instances extracted from complex indoor scenes. We adopt the same data split protocol as CLIP$^2$~\cite{zeng2023clip2}, which contains 17 semantic categories.
As shown in~\autoref{tab:zsr-realworld}, \methodsname with 640 dimensions ($d = 640$) achieves the highest overall performance with a mean Top-1 accuracy of 47.0\%, outperforming OpenShape~\cite{liu2023openshape} under the evaluated protocol by $+1.4\%$ and $\text{CLIP}^2$~\cite{zeng2023clip2} by $+8.5\%$. 
Notably, these improvements are achieved with a substantially more compact embedding than OpenShape ($d = 640$ vs. $\mathcal{D} = 1280$), highlighting the effectiveness of the proposed multi-granularity representation.

\begin{table*}[h]
\centering
\small
\setlength{\tabcolsep}{4pt}
\resizebox{\textwidth}{!}{%
\begin{tabular}{c|c|c|ccccccccccccccccc}
\toprule
Method & $\mathcal{D}$ & Avg. & Bed & Cab & Chair & Sofa & Tabl & Door & Wind & Bksf & Pic & Cntr & Desk & Curt & Fridg & Bath & Showr & Toil & Sink \\
\midrule

CLIP2Point~\cite{huang2023clip2point} & -- & 24.9 & 20.8 & 0.0 & \textbf{85.1} & 43.3 & 26.5 & \textbf{69.9} & 0.0 & 20.9 & 1.7 & 31.7 & 27.0 & 0.0 & 1.6 & 46.5 & 0.0 & 22.4 & 25.6 \\

PointCLIP w/ TP.~\cite{zhang2022pointclip} & -- & 26.1 & 0.0 & 55.7 & 72.8 & 5.0 & 5.1 & 1.7 & 0.0 & \textbf{77.2} & 0.0 & 0.0 & 51.7 & 0.3 & 0.0 & 0.0 & 40.3 & 85.3 & 49.2 \\

CLIP2Point w/ TP.~\cite{huang2023clip2point} & -- & 35.2 & 11.8 & 3.0 & 45.1 & 27.6 & 10.5 & 61.5 & 2.6 & 71.9 & 0.3 & 33.6 & 29.9 & 4.7 & 11.5 & \textbf{72.2} & \textbf{92.4} & \textbf{86.1} & 34.0 \\

CLIP$^2$~\cite{zeng2023clip2} & -- & 38.5 & 32.6 & 67.2 & 69.3 & 42.3 & 18.3 & 19.1 & 4.0 & 62.6 & 1.4 & 12.7 & 52.8 & 40.1 & 9.1 & 59.7 & 41.0 & 71.0 & 45.5 \\

OpenShape$^{\dagger}$~\cite{liu2023openshape} & 1280 & 45.6 & \textbf{66.7} & 3.2 & 75.8 & 83.5 & 37.7 & 49.2 & \textbf{47.5} & 64.9 & \textbf{48.2} & 1.9 & 66.1 & \textbf{70.2} & 1.8 & 50.0 & 57.1 & 45.2 & 7.1 \\
\midrule

\methodsname$^{\dagger}$ (Ours) & 10 & 5.7 & 10.4 & 16.7 & 3.9 & 0.0 & 0.3 & 0.2 & 0.0 & 31.6 & 0.0 & 0.0 & 0.8 & 0.0 & 0.8 & 0.0 & 29.6 & 0.0 & 2.0 \\

\methodsname$^{\dagger}$ (Ours) & 20 & 10.0 & 36.4 & 14.0 & 28.7 & 3.1 & 0.9 & 1.5 & 0.4 & 57.9 & 11.9 & 1.9 & 6.3 & 0.0 & 2.7 & 0.0 & 3.7 & 0.0 & 1.0 \\

\methodsname$^{\dagger}$ (Ours) & 40 & 11.8 & 36.4 & 0.5 & 28.7 & 3.1 & 3.1 & 16.3 & 8.5 & \textbf{77.2} & 3.0 & 5.8 & 1.6 & 0.0 & 4.6 & 0.0 & 11.1 & 0.0 & 0.0 \\

\methodsname$^{\dagger}$ (Ours) & 80 & 19.5 & 33.8 & 4.5 & 26.0 & 10.3 & 8.6 & 4.9 & 34.0 & 47.4 & 3.0 & 15.4 & 0.8 & 0.0 & 0.0 & 0.0 & 85.2 & 10.2 & 48.0 \\

\methodsname$^{\dagger}$ (Ours) & 160 & 27.4 & 35.1 & 2.3 & 15.9 & 77.3 & 38.9 & 3.9 & 17.0 & 73.7 & 22.4 & 11.5 & 1.6 & 22.6 & 0.0 & 17.3 & 70.4 & 16.9 & 38.8 \\

\methodsname$^{\dagger}$ (Ours) & 320 & 38.4 & 55.8 & 68.5 & 80.1 & 50.5 & 42.0 & 40.7 & 14.2 & 56.1 & 25.4 & 19.2 & 48.8 & 3.2 & 0.0 & 40.7 & 11.1 & 37.3 & \textbf{59.2} \\

\methodsname$^{\dagger}$ (Ours) & 640 & \textbf{47.0} & 66.2 & \textbf{71.2} & 70.8 & \textbf{84.5} & 39.7 & 32.5 & 34.4 & 38.6 & 22.4 & 57.7 & \textbf{71.2} & 67.7 & 0.3 & 53.1 & 29.6 & 47.5 & 40.8 \\

\methodsname$^{\dagger}$ (Ours) & 1280 & 40.6 & 64.9 & 64.9 & 66.8 & 79.4 & \textbf{48.8} & 33.8 & 24.1 & 15.8 & 34.3 & \textbf{61.3} & 48.8 & 13.5 & \textbf{38.6} & 49.4 & 18.5 & 44.1 & 41.8 \\

\bottomrule
\end{tabular}
}
\scriptsize
w/ TP. denotes training with the real-world data provided by CLIP$^2$. $^{\dagger}$ results using PointBERT~\cite{devlin2019bert} as 3D encoder, pre-trained on the Ensembled dataset. Avg.: Mean average Top-1 accuracy across all categories.

\vspace{-2pt} 

\caption{Zero-shot recognition on the real-world ScanNet~\cite{dai2017scannet} dataset. Bold values indicate the best performance for each category. \methodsname achieves the best results across the granularity dimensions. Bold values indicate the best results per category.}
\label{tab:zsr-realworld}
\end{table*}

We also observe progressive improvement in recognition performance as embedding dimensionality increases from 10 to 640 dimensions, as reflected in the mean top-1 accuracy, indicating that higher-dimensional subspaces progressively recover more geometric details and semantic information. In this case, the best performance is achieved at 640 dimensions rather than at 1280. This suggests that intermediate-dimensional embeddings provide a better balance between semantic abstraction and geometric specialization in real-world scenarios with noise, occlusions, and partial observations. Furthermore, different object categories achieve their best performance at different granularity levels, further suggesting that distinct semantic concepts may benefit from different levels of representational specificity.
Despite using only 320 dimensions,~\methodsname achieves a competitive mean average top-1 accuracy of 38.4\%, demonstrating that a substantial slice of the semantic knowledge learned during the multimodal pre-training is preserved even within compact embeddings.

\subsection{Cross-Modal Granularity Applications}
\label{subsec:cross-modal}

The nested structure learned by \methodsname naturally supports coarse-to-fine retrieval across modalities. Given a point cloud, image, or text query, we retrieve 3D shapes from Objaverse-LVIS~\cite{deitke2023objaverse} using cosine similarity and FAISS~\cite {johnson2019faiss}-based nearest-neighbor search. As the embedding dimensionality varies, the retrieved results shift from broad semantic categories at low dimensions to increasingly fine-grained geometric and semantic matches at higher dimensions. This behavior emerges from a single trained model and highlights the hierarchical organization of information within the learned representation space.

\begin{figure}[h]
    \centering
    \includegraphics[width=\linewidth]{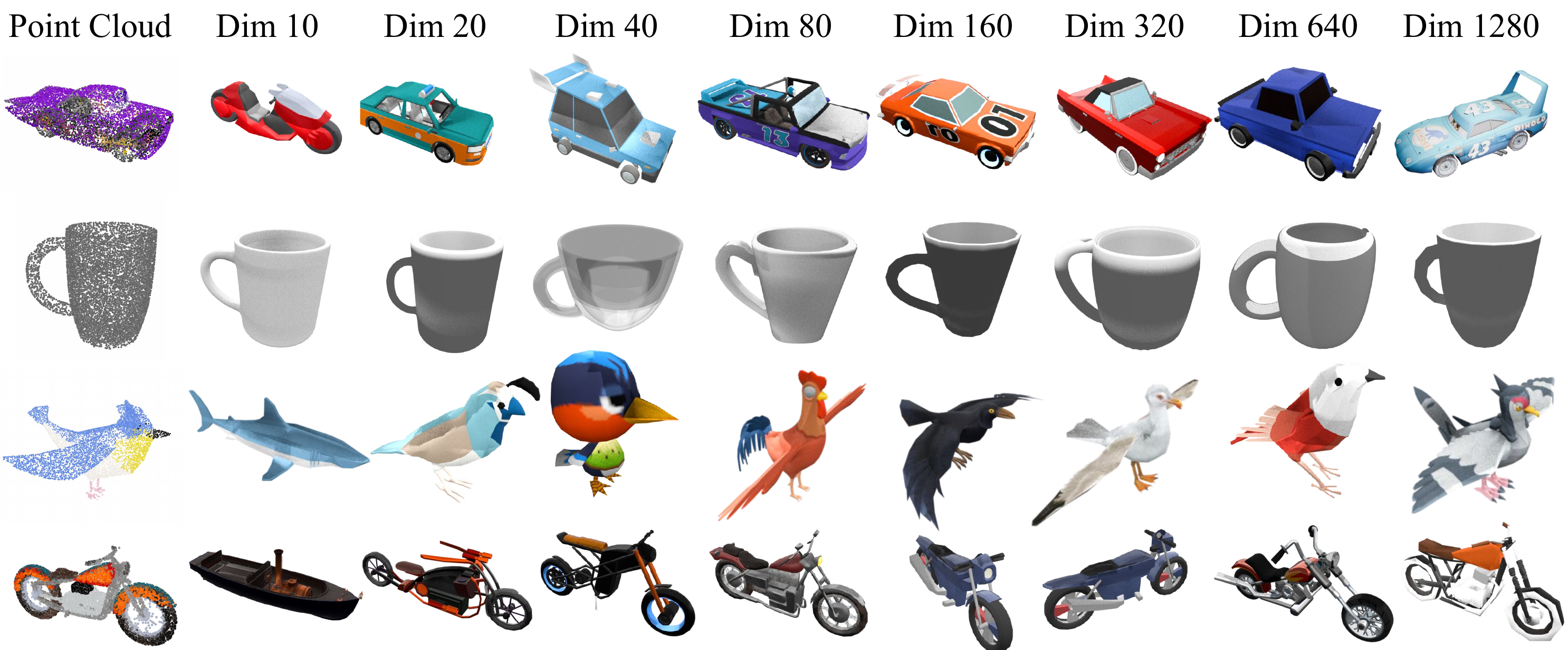}
    \caption{Point cloud-to-3D shape retrieval results using the proposed \methodsname embedding space across increasing embedding dimensions. Starting from coarse semantic retrieval at low dimensions, the representations progressively refine toward fine-grained geometric similarity at higher dimensions, demonstrating the hierarchical Matryoshka property of the learned embedding space.}
    \label{fig:3dto3d}
\end{figure}

\vspace{0.15cm}
\noindent \textbf{Point cloud-input 3D shape retrieval.}
~\autoref{fig:3dto3d} shows the qualitative behavior of the proposed~\methodsname framework in a cross-modal 3D retrieval setting.
Given a point cloud query (leftmost column), we retrieve the nearest 3D shapes from the Objaverse-LVIS~\cite{deitke2023objaverse} dataset using nested embeddings of increasing dimensionality, ranging from log-based dimensions. The results demonstrate a clear coarse-to-fine semantic hierarchy across the embedding space.

\begin{wrapfigure}{r}{0.45\textwidth}
    \vspace{-0.5cm}
    \includegraphics[width=\linewidth]{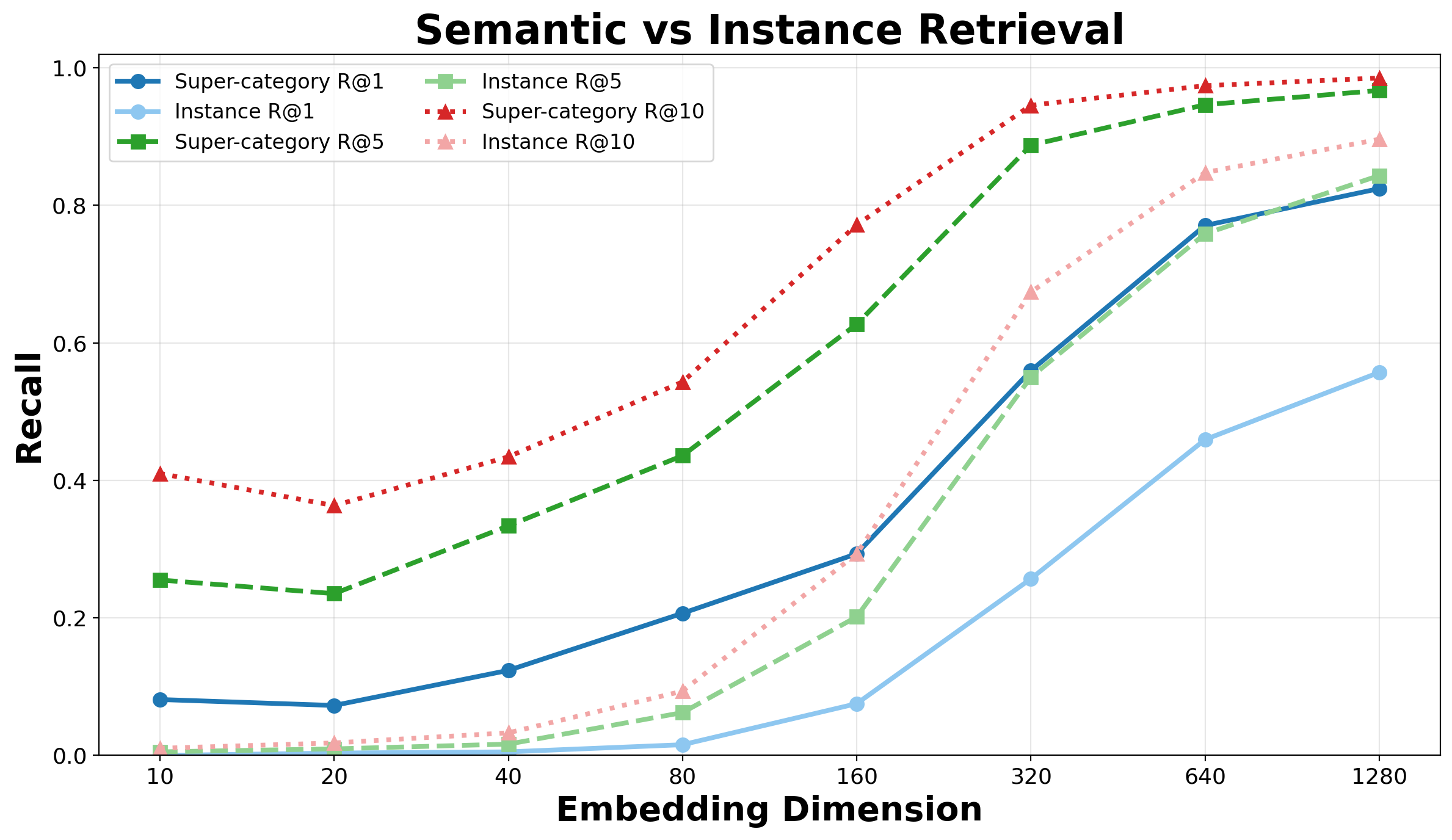}
    \vspace{-0.7cm}
    \caption{Semantic-vs-instance retrieval across embedding dimensions on Objaverse-LVIS. 
    }
    \label{fig:semantic-instance}
\end{wrapfigure}

In the first row, the query corresponds to a vehicle object. At very low dimensions (Dim 10 and Dim 20), the retrieved objects preserve only broad semantic information, yielding generic transportation-related shapes such as motorcycles and cars. As the embedding dimensionality increases, the retrieved objects progressively become more geometrically aligned with the query, converging to vehicles with similar proportions and structure at higher dimensions. The remaining rows show similar behavior. 
The motorcycle object (fourth row) shows that low-dimensional embeddings retrieve broad vehicle categories, while higher-dimensional embeddings increasingly specialize toward motorcycles with similar frame geometry, wheel configuration, and color.

\noindent\textbf{Semantic-vs-instance retrieval.}
To quantitatively assess how retrieval specificity changes across embedding dimensions, we additionally evaluate retrieval at semantic super-category and instance/category levels on Objaverse-LVIS in~\autoref{fig:semantic-instance}. 
Lower-dimensional prefixes already recover broad semantic relationships, whereas increasing embedding dimensionality progressively improves retrieval at more specific levels. 
This trend complements the qualitative results below and provides quantitative evidence for the coarse-to-fine organization of the nested representation. \emph{Complete Recall@K, mAP@100, and nDCG@10 results are reported in the supplementary material}.

\begin{figure}[h]
    \centering
    \includegraphics[width=\linewidth]{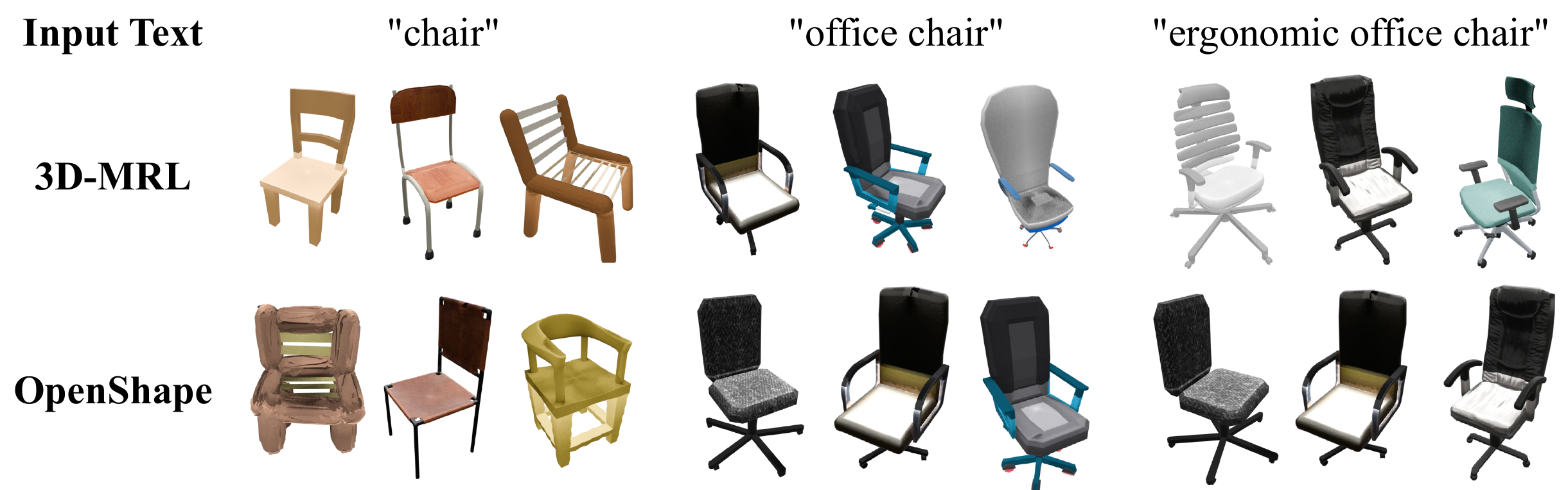}
    \caption{Text-to-3D shape retrieval comparisons.
    Qualitative text-to-3D retrieval comparisons on Objaverse-LVIS. Given progressively more specific text prompts, \methodsname produces more fine-grained and semantically accurate 3D retrievals compared to OpenShape. We list the input text and the first three retrieved 3D objects.}
    \label{fig:text3d}
\end{figure}

\vspace{0.15cm}
\noindent \textbf{Text-input 3D shape retrieval.}
In~\autoref{fig:text3d}, we show that \methodsname supports retrieving 3D shapes from detailed text descriptions, which include coarse-to-fine subcategories, attributes, and their combinations, in comparison with OpenShape~\cite{liu2023openshape}. 
We can observe this under progressively more specific text prompts. This behavior shows that \methodsname yields more semantically aligned coarse-to-fine retrieval than OpenShape at the same dimensionality (e.g., 1280).
The most challenging query, ``ergonomic office chair'', further highlights the proposed representation's fine-grained retrieval capability. \methodsname successfully retrieves highly relevant office chairs with ergonomic designs, including curved backrests, cushioned seating, and adjustable structures. Meanwhile, OpenShape retrieves objects that partially match the semantic concept but fail to capture the full ergonomic, structural, and geometric characteristics described in the text prompt.
Note that these input texts are typically not present in the raw text of retrieved shapes, indicating that~\methodsname can effectively learn generalizable concepts across the last refined dimension. \emph{Please refer to the supplementary material for more results and details}.

\vspace{-0.8cm}
\subsection{Ablation Study}
\label{subsec:ablations}

In this section, we conduct ablation experiments to validate our architectural and hyperparameter choices for \methodsname. Specifically, we investigate: (i) the effect of weight-tying in multi-scale contrastive optimization (i.e., Efficient vs. Non-Efficient mode), and (ii) the configuration of the nesting dimension schedule. Due to limited computation, we pre-train PointBERT~\cite{yu2022point} on the small-scale ShapeNet~\cite{chang2015shapenet} dataset and use two more challenging zero-shot classification benchmarks, Objaverse-LVIS~\cite{deitke2023objaverse} and ModelNet-40~\cite{wu20153dmodelnet} for evaluation.~\emph{We provide additional ablations on the number of nesting scales and modality alignment in the supplementary material}.

\vspace{0.15cm}
\noindent \textbf{Impact of Efficient MRL Formulations.} 
We first evaluate the impact of utilizing a para\-meter-free slicing layout versus employing independent classification heads during pre-train\-ing. In the supervised setting, Efficient MRL (MRL-E) reduces memory costs by tying classifier weights:$\mathbf{W}^{(m)} = \mathbf{W}_{:,1:m}$, where $\mathbf{W}_{:,1:m}$ denotes the first $m$ columns of a shared classifier matrix.
In our multimodal contrastive adaptation, the non-efficient variant (Efficient $=$ False) maintains distinct scale-specific projection matrices
$
\left\{\mathbf{W}^{(d_k)}\right\}_{k=1}^{K},
\mathbf{W}^{(d_k)} \in \mathbb{R}^{d_k \times d_k},
$
where each projection operates on the truncated representation of dimension $d_k$.
By contrast, the efficient variant (Efficient = True) removes scale-specific projection heads and instead performs parameter-free geometric truncation, thereby sharing projection parameters across scales.

As shown in~\autoref{tab:eff_mode}, the non-efficient configuration yields a minor, near-negligible performance edge on Objaverse-LVIS (22.1\% vs.\ 21.6\% Top-3 accuracy) and ModelNet-40 (71.8\% vs.\ 71.5\% Top-1 accuracy). However, this marginal gain comes at the cost of maintaining $K$ distinct parameter matrices during joint contrastive alignment. By enforcing parameter-free geometric truncation, the efficient configuration substantially reduces parameter overhead and avoids optimization bottlenecks, while maintaining highly competitive representational capacity across all benchmarks.
\begin{table}[htpb]

\centering
\begin{tabular}{c | c c c | c c c}
\multirow{2}{*}{Efficient} & \multicolumn{3}{c|}{Objaverse-LVIS} & \multicolumn{3}{c}{ModelNet-40} \\ & Top-1 & Top-3 & Top-5 & Top-1 & Top-3 & Top-5 \\
\hline
True & 12.4 & 21.6 & 26.5 & 71.5 & \textbf{87.1} & 91.1 \\
\rowcolor{Gray!10} False & \textbf{12.4} & \textbf{22.1} & \textbf{27.2} & \textbf{71.8} & 86.8 & \textbf{91.3} \\
\end{tabular}
\vspace{5pt}
\caption{Effect of weight-tying: Efficient vs. Non-Efficient mode. We compare the parameter-free geometric truncation (True) against the use of independent scale-specific linear projection heads (False). In bold, we display the best performance per column.} 
\label{tab:eff_mode}
\end{table}

\vspace{0.15cm}
\noindent \textbf{Evaluation of Nesting Dimension Schedules.} 
The choice of the dimension schedule $\mathcal{D}$ determines how representational capacity is distributed across the coarse-to-fine spectrum. We ablate three distinct schedules targeting a maximum capacity of $\mathcal{D}=1280$:
\begin{itemize}
    \item \textbf{Full:} A standard baseline optimizing solely for the maximum fixed-dimensional embedding space ($\mathcal{D}=1280$).
    \item \textbf{Linear:} A uniform distribution spanning nine dimensions from base to ceiling ($\mathcal{D} = \{8, 16, 32, \dots, 1280\}$).
    \item \textbf{Log-based:} Our proposed geometric progression starting at a base dimension of 10 and doubling at each step ($\mathcal{D} = \{10, 20, 40, \dots, 1280\}$).
\end{itemize}

The empirical comparisons compiled in~\autoref{tab:schedule_train} reveal that the \emph{log-based} schedule consistently outperforms both the linear and full-dimensional baselines. Specifically, on ModelNet40, the log-based distribution achieves $71.8\%$ Top-1 accuracy, outperforming the full baseline by $+1.2\%$ and the linear schedule by $+2.0\%$. On Objaverse-LVIS, the log-based approach matches or exceeds alternative schedules across all metrics (e.g., $12.4\%$ Top-1 and $27.2\%$ Top-5 accuracy).

These results indicate that allocating representational granularities exponentially (i.e., $O(\log(\mathcal{D}))$ scales) provides a superior optimization regularizer compared to standard uniform schedules. By densely supervising the lower-dimensional spaces (where representational capacity scales most critically) while smoothly interpolating toward higher dimensions, the log-based schedule prevents over-parameterization and induces a significantly more robust multimodal alignment.
\begin{table}[htpb]
\centering

\resizebox{\textwidth}{!}{%
\begin{tabular}{c | c | c c c | c c c}
\multirow{2}{*}{\begin{tabular}{c}
Embedding \\ Schedule
\end{tabular}} &
\multirow{2}{*}{Nesting dimensions} &
\multicolumn{3}{c|}{Objaverse-LVIS} &
\multicolumn{3}{c}{ModelNet-40} \\

& &
Top-1 & Top-3 & Top-5 &
Top-1 & Top-3 & Top-5 \\
\hline

full &
1280 &
  12.1 & 22.1 & 27.0 & 70.6 & 86.0 & 90.3
   \\

linear &
8, 16, 32, 64, 128, 256, 512, 1024, 1280 &
 12.4 & 22.1 & 27.2 & 69.8 & 86.1 & 91.3  \\

\rowcolor{gray!10} log-based &
10, 20, 40, 80, 160, 320, 640, 1280 &
 \textbf{12.4} & \textbf{22.1} & \textbf{27.2} & \textbf{71.8} & \textbf{86.8} & \textbf{91.3} \\

\end{tabular}
}
\vspace{5pt}
\caption{Embedding dimension schedules. Comparison of zero-shot classification performance across different distribution granularities for a maximum capacity of $\mathcal{D}=1280$. Our proposed geometric \emph{log-based} schedule consistently outperforms both the fixed-dimensional baseline (\emph{full}) and the uniform sequence (\emph{linear}), serving as an effective multi-scale optimization regularizer.  In bold, we display the best performance per column.}
\label{tab:schedule_train}

\end{table}

\vspace{-0.9cm}
\section{Conclusion}
\label{sec:conclusion}
This paper introduced \methodsname, the first framework to bring Matryoshka Representation Learning to multimodal 3D pre-training. By learning nested 3D representations aligned with frozen image and text embeddings, \methodsname enables a single model to operate across multiple dimensionality budgets without retraining. Experiments on Objaverse-LVIS, ModelNet40, ScanObjectNN, and ScanNet show that the proposed framework achieves competitive performance while consistently improving the accuracy-efficiency trade-off of multimodal 3D representations.
Our experiments further suggest that the nested structure learned by \methodsname organizes information hierarchically across embedding dimensions, enabling retrieval behavior that progressively transitions from broad semantic similarity to finer geometric and semantic specificity. These results highlight the potential of nested representations as a practical foundation for scalable multimodal 3D understanding, retrieval, and deployment under diverse computational constraints.

\vspace{-0.5cm}
\section{Acknowledgments}

This work was supported in part by the São Paulo Research Foundation (FAPESP), under grants \#2024/09462-1 and \#2026/01721-3, and by the Conselho Nacional de Desenvolvimento Científico e Tecnológico (CNPq) under fellowship grant \#315158/2023-9. The authors gratefully acknowledge the Center for Mathematical Sciences Applied to Industry (CeMEAI) for providing computational resources, funded by FAPESP under grant \#2013/07375-0.

\vspace{-0.6cm}
\bibliography{egbib}

@article{liu2023openshape,
  title={Openshape: Scaling up 3d shape representation towards open-world understanding},
  author={Liu, Minghua and Shi, Ruoxi and Kuang, Kaiming and Zhu, Yinhao and Li, Xuanlin and Han, Shizhong and Cai, Hong and Porikli, Fatih and Su, Hao},
  journal={Advances in neural information processing systems},
  volume={36},
  pages={44860--44879},
  year={2023}
}

@inproceedings{zhang2024tamm,
  title={Tamm: Triadapter multi-modal learning for 3d shape understanding},
  author={Zhang, Zhihao and Cao, Shengcao and Wang, Yu-Xiong},
  booktitle={Proceedings of the IEEE/CVF conference on computer vision and pattern recognition},
  pages={21413--21423},
  year={2024}
}

@inproceedings{gao2024sculpting,
  title={Sculpting holistic 3d representation in contrastive language-image-3d pre-training},
  author={Gao, Yipeng and Wang, Zeyu and Zheng, Wei-Shi and Xie, Cihang and Zhou, Yuyin},
  booktitle={Proceedings of the IEEE/CVF Conference on Computer Vision and Pattern Recognition},
  pages={22998--23008},
  year={2024}
}

@inproceedings{zhou2024uni3d,
  title={Uni3d: Exploring unified 3d representation at scale},
  author={Zhou, Junsheng and Wang, Jinsheng and Ma, Baorui and Liu, Yu-Shen and Huang, Tiejun and Wang, Xinlong},
  booktitle={International Conference on Learning Representations},
  volume={2024},
  pages={46766--46782},
  year={2024}
}

@article{chang2015shapenet,
  title={Shapenet: An information-rich 3d model repository},
  author={Chang, Angel X and Funkhouser, Thomas and Guibas, Leonidas and Hanrahan, Pat and Huang, Qixing and Li, Zimo and Savarese, Silvio and Savva, Manolis and Song, Shuran and Su, Hao and others},
  journal={arXiv preprint arXiv:1512.03012},
  year={2015}
}

@article{future20213d,
  title={3d-future: 3d furniture shape with texture},
  author={Fu, Huan and Jia, Rongfei and Gao, Lin and Gong, Mingming and Zhao, Binqiang and Maybank, Steve and Tao, Dacheng},
  journal={International Journal of Computer Vision},
  volume={129},
  number={12},
  pages={3313--3337},
  year={2021},
  publisher={Springer}
}

@inproceedings{collins2022abo,
  title={Abo: Dataset and benchmarks for real-world 3d object understanding},
  author={Collins, Jasmine and Goel, Shubham and Deng, Kenan and Luthra, Achleshwar and Xu, Leon and Gundogdu, Erhan and Zhang, Xi and Vicente, Tomas F Yago and Dideriksen, Thomas and Arora, Himanshu and others},
  booktitle={Proceedings of the IEEE/CVF conference on computer vision and pattern recognition},
  pages={21126--21136},
  year={2022}
}

@inproceedings{lee2025duoduo,
  title={Duoduo CLIP: Efficient 3D understanding with multi-view images},
  author={Lee, Han-Hung and Zhang, Yiming and Chang, Angel},
  booktitle={International Conference on Learning Representations},
  volume={2025},
  pages={48070--48091},
  year={2025}
}

@article{oord2018representation,
  title={Representation learning with contrastive predictive coding},
  author={Oord, Aaron van den and Li, Yazhe and Vinyals, Oriol},
  journal={arXiv preprint arXiv:1807.03748},
  year={2018}
}

@inproceedings{zhou2025crossmodal,
  title={Cross-modal 3d representation with multi-view images and point clouds},
  author={Zhou, Ziyang and Wang, Pinghui and Liang, Zi and Bai, Haitao and Zhang, Ruofei},
  booktitle={Proceedings of the Computer Vision and Pattern Recognition Conference},
  pages={3728--3739},
  year={2025}
}

@article{song2025adaptive,
  title={Adaptive CLIP for open-domain 3D model retrieval},
  author={Song, Dan and Qiang, Zekai and Zhang, Chumeng and Wang, Lanjun and Liu, Qiong and Yang, You and Liu, An-An},
  journal={Information Processing \& Management},
  volume={62},
  number={2},
  pages={103989},
  year={2025},
  publisher={Elsevier}
}

@inproceedings{deitke2023objaverse,
  title={Objaverse: A universe of annotated 3d objects},
  author={Deitke, Matt and Schwenk, Dustin and Salvador, Jordi and Weihs, Luca and Michel, Oscar and VanderBilt, Eli and Schmidt, Ludwig and Ehsani, Kiana and Kembhavi, Aniruddha and Farhadi, Ali},
  booktitle={Proceedings of the IEEE/CVF conference on computer vision and pattern recognition},
  pages={13142--13153},
  year={2023}
}

@inproceedings{li2022blip,
  title={Blip: Bootstrapping language-image pre-training for unified vision-language understanding and generation},
  author={Li, Junnan and Li, Dongxu and Xiong, Caiming and Hoi, Steven},
  booktitle={International conference on machine learning},
  pages={12888--12900},
  year={2022},
  organization={PMLR}
}

@inproceedings{li2023blip,
  title={Blip-2: Bootstrapping language-image pre-training with frozen image encoders and large language models},
  author={Li, Junnan and Li, Dongxu and Savarese, Silvio and Hoi, Steven},
  booktitle={International conference on machine learning},
  pages={19730--19742},
  year={2023},
  organization={PMLR}
}

@article{achiam2023gpt,
  title={Gpt-4 technical report},
  author={Achiam, Josh and Adler, Steven and Agarwal, Sandhini and Ahmad, Lama and Akkaya, Ilge and Aleman, Florencia Leoni and Almeida, Diogo and Altenschmidt, Janko and Altman, Sam and Anadkat, Shyamal and others},
  journal={arXiv preprint arXiv:2303.08774},
  year={2023}
}

@inproceedings{gupta2019lvis,
  title={Lvis: A dataset for large vocabulary instance segmentation},
  author={Gupta, Agrim and Dollar, Piotr and Girshick, Ross},
  booktitle={Proceedings of the IEEE/CVF conference on computer vision and pattern recognition},
  pages={5356--5364},
  year={2019}
}

@inproceedings{wu20153dmodelnet,
  title={3d shapenets: A deep representation for volumetric shapes},
  author={Wu, Zhirong and Song, Shuran and Khosla, Aditya and Yu, Fisher and Zhang, Linguang and Tang, Xiaoou and Xiao, Jianxiong},
  booktitle={Proceedings of the IEEE conference on computer vision and pattern recognition},
  pages={1912--1920},
  year={2015}
}

@inproceedings{scanobj2019,
  title={Revisiting point cloud classification: A new benchmark dataset and classification model on real-world data},
  author={Uy, Mikaela Angelina and Pham, Quang-Hieu and Hua, Binh-Son and Nguyen, Thanh and Yeung, Sai-Kit},
  booktitle={Proceedings of the IEEE/CVF international conference on computer vision},
  pages={1588--1597},
  year={2019}
}

@inproceedings{xue2023ulip,
  title={Ulip: Learning a unified representation of language, images, and point clouds for 3d understanding},
  author={Xue, Le and Gao, Mingfei and Xing, Chen and Mart{\'\i}n-Mart{\'\i}n, Roberto and Wu, Jiajun and Xiong, Caiming and Xu, Ran and Niebles, Juan Carlos and Savarese, Silvio},
  booktitle={Proceedings of the IEEE/CVF conference on computer vision and pattern recognition},
  pages={1179--1189},
  year={2023}
}

@inproceedings{choy20194dsparseconv,
  title={4d spatio-temporal convnets: Minkowski convolutional neural networks},
  author={Choy, Christopher and Gwak, JunYoung and Savarese, Silvio},
  booktitle={Proceedings of the IEEE/CVF conference on computer vision and pattern recognition},
  pages={3075--3084},
  year={2019}
}

@article{kusupati2022matryoshka,
  title={Matryoshka representation learning},
  author={Kusupati, Aditya and Bhatt, Gantavya and Rege, Aniket and Wallingford, Matthew and Sinha, Aditya and Ramanujan, Vivek and Howard-Snyder, William and Chen, Kaifeng and Kakade, Sham and Jain, Prateek and others},
  journal={Advances in Neural Information Processing Systems},
  volume={35},
  pages={30233--30249},
  year={2022}
}

@article{li20242dmrl,
  title={2d matryoshka sentence embeddings},
  author={Li, Xianming and Li, Zongxi and Li, Jing and Xie, Haoran and Li, Qing},
  journal={arXiv preprint arXiv:2402.14776},
  year={2024}
}

@inproceedings{yu2022point,
  title={Point-bert: Pre-training 3d point cloud transformers with masked point modeling},
  author={Yu, Xumin and Tang, Lulu and Rao, Yongming and Huang, Tiejun and Zhou, Jie and Lu, Jiwen},
  booktitle={Proceedings of the IEEE/CVF conference on computer vision and pattern recognition},
  pages={19313--19322},
  year={2022}
}

@article{zhang2022point,
  title={Point-m2ae: multi-scale masked autoencoders for hierarchical point cloud pre-training},
  author={Zhang, Renrui and Guo, Ziyu and Gao, Peng and Fang, Rongyao and Zhao, Bin and Wang, Dong and Qiao, Yu and Li, Hongsheng},
  journal={Advances in neural information processing systems},
  volume={35},
  pages={27061--27074},
  year={2022}
}

@inproceedings{zhang2022pointclip,
  title={Pointclip: Point cloud understanding by clip},
  author={Zhang, Renrui and Guo, Ziyu and Zhang, Wei and Li, Kunchang and Miao, Xupeng and Cui, Bin and Qiao, Yu and Gao, Peng and Li, Hongsheng},
  booktitle={Proceedings of the IEEE/CVF conference on computer vision and pattern recognition},
  pages={8552--8562},
  year={2022}
}

@inproceedings{zhu2023pointclipv2,
  title={Pointclip v2: Prompting clip and gpt for powerful 3d open-world learning},
  author={Zhu, Xiangyang and Zhang, Renrui and He, Bowei and Guo, Ziyu and Zeng, Ziyao and Qin, Zipeng and Zhang, Shanghang and Gao, Peng},
  booktitle={Proceedings of the IEEE/CVF international conference on computer vision},
  pages={2639--2650},
  year={2023}
}

@inproceedings{huang2023clip2point,
  title={Clip2point: Transfer clip to point cloud classification with image-depth pre-training},
  author={Huang, Tianyu and Dong, Bowen and Yang, Yunhan and Huang, Xiaoshui and Lau, Rynson WH and Ouyang, Wanli and Zuo, Wangmeng},
  booktitle={Proceedings of the IEEE/CVF international conference on computer vision},
  pages={22157--22167},
  year={2023}
}

@inproceedings{qi2023contrast,
  title={Contrast with reconstruct: Contrastive 3d representation learning guided by generative pretraining},
  author={Qi, Zekun and Dong, Runpei and Fan, Guofan and Ge, Zheng and Zhang, Xiangyu and Ma, Kaisheng and Yi, Li},
  booktitle={International Conference on Machine Learning},
  pages={28223--28243},
  year={2023},
  organization={PMLR}
}

@article{xiao2025metaembed,
  title={Metaembed: Scaling multimodal retrieval at test-time with flexible late interaction},
  author={Xiao, Zilin and Ma, Qi and Gu, Mengting and Chen, Chun-cheng Jason and Chen, Xintao and Ordonez, Vicente and Mohan, Vijai},
  journal={arXiv preprint arXiv:2509.18095},
  year={2025}
}

@inproceedings{radford2021learning,
  title={Learning transferable visual models from natural language supervision},
  author={Radford, Alec and Kim, Jong Wook and Hallacy, Chris and Ramesh, Aditya and Goh, Gabriel and Agarwal, Sandhini and Sastry, Girish and Askell, Amanda and Mishkin, Pamela and Clark, Jack and others},
  booktitle={International conference on machine learning},
  pages={8748--8763},
  year={2021},
  organization={PmLR}
}

@inproceedings{cherti2023reproducible,
  title={Reproducible scaling laws for contrastive language-image learning},
  author={Cherti, Mehdi and Beaumont, Romain and Wightman, Ross and Wortsman, Mitchell and Ilharco, Gabriel and Gordon, Cade and Schuhmann, Christoph and Schmidt, Ludwig and Jitsev, Jenia},
  booktitle={Proceedings of the IEEE/CVF conference on computer vision and pattern recognition},
  pages={2818--2829},
  year={2023}
}

@inproceedings{qi2017pointnet,
  title={Pointnet: Deep learning on point sets for 3d classification and segmentation},
  author={Qi, Charles R and Su, Hao and Mo, Kaichun and Guibas, Leonidas J},
  booktitle={Proceedings of the IEEE conference on computer vision and pattern recognition},
  pages={652--660},
  year={2017}
}

@article{qi2017pointnet++,
  title={Pointnet++: Deep hierarchical feature learning on point sets in a metric space},
  author={Qi, Charles Ruizhongtai and Yi, Li and Su, Hao and Guibas, Leonidas J},
  journal={Advances in neural information processing systems},
  volume={30},
  year={2017}
}

@inproceedings{shi2020pv,
  title={Pv-rcnn: Point-voxel feature set abstraction for 3d object detection},
  author={Shi, Shaoshuai and Guo, Chaoxu and Jiang, Li and Wang, Zhe and Shi, Jianping and Wang, Xiaogang and Li, Hongsheng},
  booktitle={Proceedings of the IEEE/CVF conference on computer vision and pattern recognition},
  pages={10529--10538},
  year={2020}
}

@article{wu2022pointtransformer,
  title={Point transformer v2: Grouped vector attention and partition-based pooling},
  author={Wu, Xiaoyang and Lao, Yixing and Jiang, Li and Liu, Xihui and Zhao, Hengshuang},
  journal={Advances in Neural Information Processing Systems},
  volume={35},
  pages={33330--33342},
  year={2022}
}

@article{zhang2022pointmae,
  title={Point-m2ae: multi-scale masked autoencoders for hierarchical point cloud pre-training},
  author={Zhang, Renrui and Guo, Ziyu and Gao, Peng and Fang, Rongyao and Zhao, Bin and Wang, Dong and Qiao, Yu and Li, Hongsheng},
  journal={Advances in neural information processing systems},
  volume={35},
  pages={27061--27074},
  year={2022}
}

@inproceedings{devlin2019bert,
  title={Bert: Pre-training of deep bidirectional transformers for language understanding},
  author={Devlin, Jacob and Chang, Ming-Wei and Lee, Kenton and Toutanova, Kristina},
  booktitle={Proceedings of the 2019 conference of the North American chapter of the association for computational linguistics: human language technologies, volume 1 (long and short papers)},
  pages={4171--4186},
  year={2019}
}

@article{guo2026mrlgs,
      title={Matryoshka Gaussian Splatting},
      author={Guo, Zhilin and Zhang, Boqiao and Aktas, Hakan and Fogarty, Kyle and Hu, Jeffrey and Aslan, Nursena Koprucu and Li, Wenzhao and Baykal, Canberk and Miao, Albert and Bengtson, Josef and Zhou, Chenliang and Xia, Weihao and Nader Vasconcelos, Cristina and Oztireli, Cengiz},
      journal={arXiv preprint arXiv:2603.19234},
      year={2026}
}

@inproceedings{cai2025matryoshka,
  title={Matryoshka multimodal models},
  author={Cai, Mu and Yang, Jianwei and Gao, Jianfeng and Lee, Yong Jae},
  booktitle={International Conference on Learning Representations},
  volume={2025},
  pages={46254--46272},
  year={2025}
}

@inproceedings{li2025triclip,
  title={TriCLIP-3D: A Unified Parameter-Efficient Framework for Tri-Modal 3D Visual Grounding based on CLIP},
  author={Li, Fan and Wang, Zanyi and Huang, Zeyi and Dai, Guang and Wang, Jingdong and Wang, Mengmeng},
  booktitle={Proceedings of the 33rd ACM International Conference on Multimedia},
  pages={5257--5266},
  year={2025}
}

@inproceedings{zeng2023clip2,
  title={Clip2: Contrastive language-image-point pretraining from real-world point cloud data},
  author={Zeng, Yihan and Jiang, Chenhan and Mao, Jiageng and Han, Jianhua and Ye, Chaoqiang and Huang, Qingqiu and Yeung, Dit-Yan and Yang, Zhen and Liang, Xiaodan and Xu, Hang},
  booktitle={Proceedings of the IEEE/CVF conference on computer vision and pattern recognition},
  pages={15244--15253},
  year={2023}
}

@inproceedings{dai2017scannet,
    title={ScanNet: Richly-annotated 3D Reconstructions of Indoor Scenes},
    author={Dai, Angela and Chang, Angel X. and Savva, Manolis and Halber, Maciej and Funkhouser, Thomas and Nie{\ss}ner, Matthias},
    booktitle = {Proc. Computer Vision and Pattern Recognition (CVPR), IEEE},
    year = {2017}
}

@article{johnson2019faiss,
  title={Billion-scale similarity search with GPUs},
  author={Johnson, Jeff and Douze, Matthijs and J{\'e}gou, Herv{\'e}},
  journal={IEEE transactions on big data},
  volume={7},
  number={3},
  pages={535--547},
  year={2019},
  publisher={IEEE}
}

@inproceedings{wang2024multimrd,
  title={Multi-modal relation distillation for unified 3d representation learning},
  author={Wang, Huiqun and Bao, Yiping and Pan, Panwang and Li, Zeming and Liu, Xiao and Yang, Ruijie and Huang, Di},
  booktitle={European Conference on Computer Vision},
  pages={364--381},
  year={2024},
  organization={Springer}
}

@inproceedings{zhang2025smec,
  title={SMEC: rethinking matryoshka representation learning for retrieval embedding compression},
  author={Zhang, Biao and Chen, Lixin and Liu, Tong and Zheng, Bo},
  booktitle={Proceedings of the 2025 Conference on Empirical Methods in Natural Language Processing},
  pages={26220--26233},
  year={2025}
}

@article{wen2025beyondcsr,
  title={Beyond matryoshka: Revisiting sparse coding for adaptive representation},
  author={Wen, Tiansheng and Wang, Yifei and Zeng, Zequn and Peng, Zhong and Su, Yudi and Liu, Xinyang and Chen, Bo and Liu, Hongwei and Jegelka, Stefanie and You, Chenyu},
  journal={arXiv preprint arXiv:2503.01776},
  year={2025}
}

@article{mao2024opendlign,
  title={Opendlign: Open-world point cloud understanding with depth-aligned images},
  author={Mao, Ye and Jing, Junpeng and Mikolajczyk, Krystian},
  journal={Advances in Neural Information Processing Systems},
  volume={37},
  pages={101144--101167},
  year={2024}
}

\end{document}